\documentclass[letterpaper, 10 pt, conference, twocolumn]{ieeeconf}  

\IEEEoverridecommandlockouts                              

\usepackage{caption}
\usepackage{hyperref}
\usepackage{graphicx}
\usepackage{amsmath} 
\usepackage{amssymb}  

\usepackage{algpseudocode}
\usepackage{algorithm}
\usepackage{booktabs}
\usepackage{multirow}

\usepackage{lipsum}
\usepackage{etoolbox}
\usepackage{caption}
\usepackage{titlesec}

\newcommand{\shrink}{\def\baselinestretch{0.99}\large\normalsize} 
\shrink

\title{\LARGE \bf Latent Policy Steering with\\Embodiment-Agnostic Pretrained World Models}

\author{Yiqi Wang$^{1}$ \thanks{$^{1}$Authors are from Robotics Institute, School of Computer Science, Carnegie Mellon University, 5000 Forbes Ave, Pittsburgh, United States {\tt\small yiqiw2@andrew.cmu.edu}}  and Mrinal Verghese$^{1}$ and Jeff Schneider$^{1}$}

\makeatletter
\apptocmd{\@maketitle}{\centering\includegraphics[width=.9\linewidth]{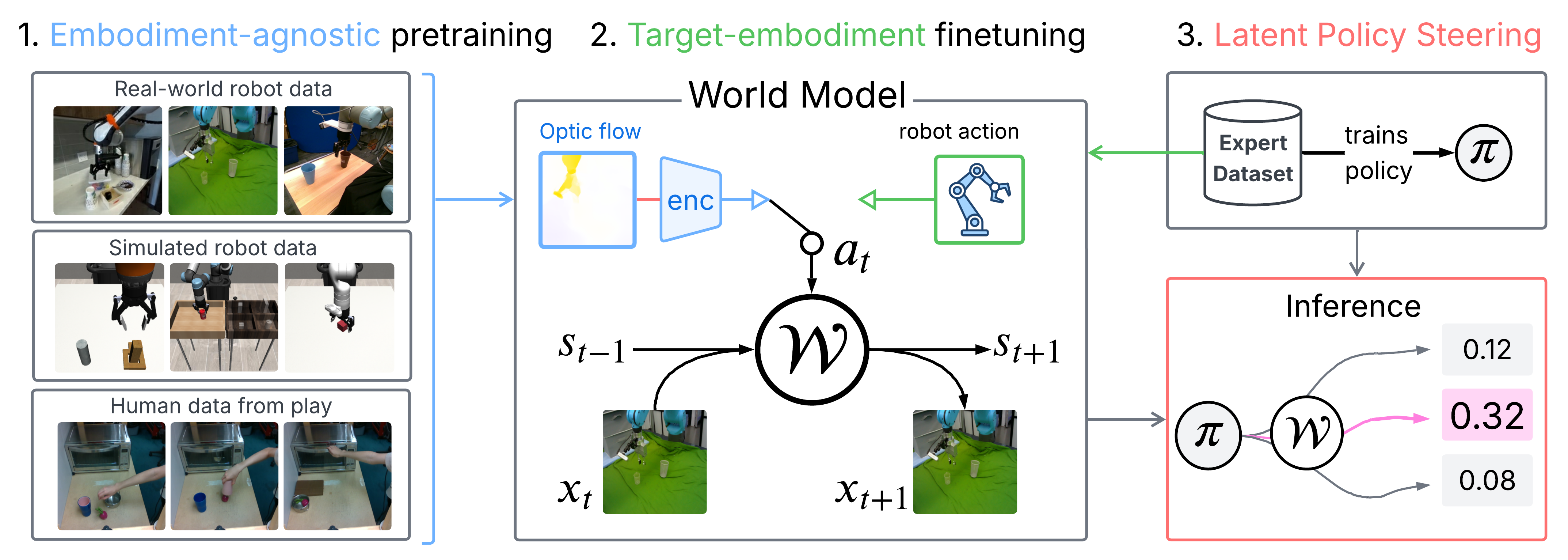}
\captionof{figure}{\textbf{Overview}. Left: We propose optical flow as an embodiment-agnostic action representation, allowing robots to make use of existing or cost-efficient data across diverse embodiments. Middle: An image-based World Model (WM) is pretrained using encoded optical flow as an action representation, and then finetuned on a target-embodiment with robot actions. Right: During inference, Latent Policy Steering evaluates multiple candidate plans and executes the best one.}\label{fig:motivation1}
\vspace{-3mm}
}{}{}

\makeatother

\begin{document}

\maketitle

\thispagestyle{empty}
\pagestyle{empty}


\begin{abstract}

  The performance of learned robot visuomotor policies is heavily dependent on the size and quality of the training dataset. Although large-scale robot and human datasets are increasingly available, embodiment gaps and mismatched action spaces make them difficult to leverage. Our main insight is that skills performed across different embodiments produce visual similarities in motions that can be captured using off-the-shelf action representations such as optical flow. Moreover, World Models (WMs) can leverage sub-optimal data since they focus on modeling dynamics.

  In this work, we aim to improve visuomotor policies in low-data regimes by first pretraining a WM using optical flow as an embodiment-agnostic action representation to leverage accessible or easily collected data from multiple embodiments (robots, humans). Given a small set of demonstrations on a target embodiment, we finetune the WM on this data to better align the WM predictions, train a base policy, and learn a robust value function. Using our finetuned WM and value function, our approach evaluates action candidates from the base policy and selects the best one to improve performance. Our approach, which we term Latent Policy Steering (LPS), improves behavior-cloned policies by 10.6\% on average across four Robomimic tasks, even though most of the pretraining data comes from the real world. In the real-world experiments, LPS achieves larger gains: 70\% relative improvement with 30-50 target-embodiment demonstrations, and 44\% relative improvement with 60-100 demonstrations, compared to a behavior-cloned baseline. Qualitative results can be found on the \href{https://yiqiwang8177.github.io/LatentPolicySteering/}{website}. 
  
\end{abstract}


\section{INTRODUCTION}

Imitation learning through Behavior Cloning (BC) is a widely adopted paradigm to acquire visuomotor policies for robots \cite{brohan2022rt}\cite{chi2023diffusion}\cite{shafiullah2022behavior}. To achieve high task success, sufficient expert demonstrations must be collected, which is a time-consuming process. Furthermore, the data collected is often specific to a robot, a task, or an environment, and the data collection process may need to be repeated for different embodiments or different environments. By collecting large datasets across different robots and environments \cite{vuong2023open}, progress has been made toward building generalist robot policies via cross-embodiment training \cite{team2024octo}\cite{henighan2020scaling}.  However, these models sometimes do not generalize to new tasks with satisfactory results. To finetune these models for better performance, a considerable amount of data is required, given the large model size used to learn from large datasets  \cite{brohan2022rt}\cite{shafiullah2022behavior}\cite{black2410pi0}. For example,  $\pi0$ \cite{black2410pi0} requires 5-10 extra hours of fine-tuning data to achieve high success rates on new tasks.

In typical LLM/VLM practice, a model trained on large and diverse datasets across multiple tasks will produce general representations that are transferable to new tasks with only a small amount of data, known as pretraining \cite{henighan2020scaling}\cite{kaplan2020scaling}. However, pretraining a robot model is challenging: each robot in the dataset has embodiment-specific information (proprioception, actions). Thus, a pretraining dataset with different embodiments makes the pretrained representation heavily dependent on the embodiments included in the dataset and may generalize poorly to new embodiments.

\begin{figure}[t]
      \centering
     
      \includegraphics[scale=0.042]{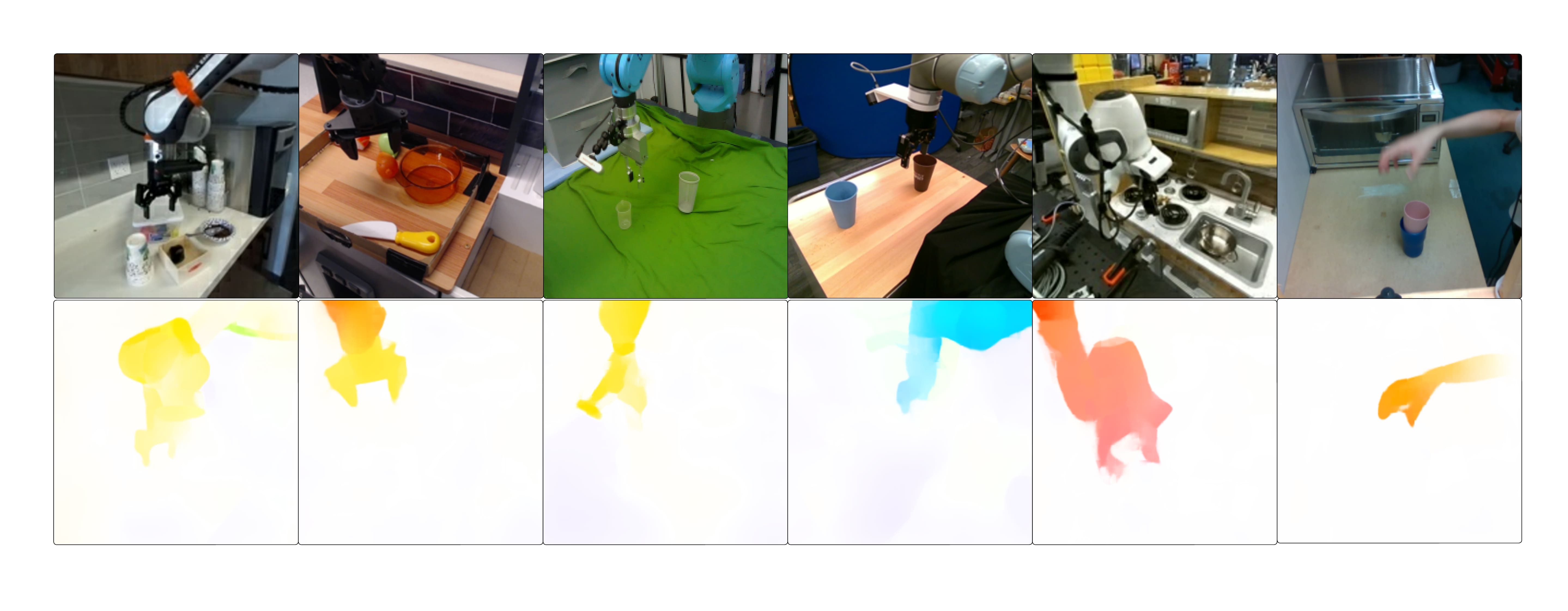}
      \caption{ 
      \textbf{Optical flow as action}. We observe that motions captured by optical flow across embodiments are similar in visual space. For example, picking up a cup will generate similar flow vectors in both the agent and the cup, regardless of agent embodiment. By using optical flow as an action representation, we reduce dependency on specific embodiments, allowing the pretrained model to more easily adapt to a new embodiment during finetuning. }
      \vspace{-1em}
      \label{fig:optical_flow_visualization}
   \end{figure}

To tackle this challenge, prior works attempt to extract shared action across embodiments through large-scale unsupervised learning \cite{schmidt2023learning}\cite{bruce2024genie}. However, we propose a more straightforward solution given our key observation: \textit{skills performed across different embodiments produce visually similar motion}, which can be captured by off-the-shelf tools such as optical flow. Optical flow allows different embodiments to share the same action space, as shown in Fig.~\ref{fig:optical_flow_visualization}. We learn an image-based WM (Dreamer \cite{hafner2019learning}\cite{hafner2023mastering}) with encoded optical flow as action (middle of the Fig.~\ref{fig:motivation1}). To encode optical flow, we train a convolution-based encoder end-to-end with the WM to produce compact encodings that suppress motion-irrelevant information (e.g., noise, morphology differences across embodiments). During pretraining, we choose to learn a WM instead of a policy, since WM learns dynamics and is more sample-efficient than a policy \cite{hafner2019learning}\cite{hafner2023mastering}.

For a given target embodiment and task, we collect a small expert dataset to learn a diffusion policy \cite{chi2023diffusion} from scratch. The WM is finetuned on the target dataset with robot actions instead of optical flow, and a value function is extracted. During the inference, the WM and value function will evaluate multiple action plans sampled from the policy, and select the best action plan for execution (bottom right of the Fig.~\ref{fig:motivation1}). This process is often referred to as Policy Steering  \cite{nakamoto2024steering}\cite{wang2024inference}\cite{wu2025foresight}.

Prior work on Policy Steering \cite{nakamoto2024steering}\cite{wang2024inference}\cite{wu2025foresight} has largely concerned steering the policy toward the user's desired outcome. In our implementation, we address a prevalent problem for sequential decision-making: distribution shift. We propose Latent Policy Steering (LPS): \textit{LPS learns a robust value function to avoid distribution shift by simulating the shift in the latent state space during finetuning}. Specifically, we use the world model in conjunction with the base policy to simulate states the policy is likely to visit during inference and train our value function across these simulated states to penalize distribution shifts from the expert data. As a result, the LPS's value function is robust to inference-time distribution shift as it has foreseen it during finetuning.

Our contributions are as follows.
\begin{itemize}
    \item We propose optical flow as an embodiment-agnostic action representation to pretrain a World Model (WM) across different robots and human embodiments. 
    \item We propose Latent Policy Steering (LPS) to align the embodiment-agnostic WM to a target-embodiment policy. To mitigate inference-time distribution shift, LPS learns a robust value function in the WM's latent space by training on both expert states and states policy is likely to visit during inference, and steers the policy back towards the expert state distribution via latent state similarities between them. 
    \item We demonstrate that LPS with a pretrained WM significantly improves the policy's performance on long-horizon manipulation in simulations, and manipulation tasks in the real-world involving tool-use and deformable objects. Our results show the effectiveness of pretraining embodiment-agnostic WMs on abundant data from other embodiments to enhance a policy in low-data scenarios.
    
\end{itemize}

\section{Related works}

\subsection{Robot Learning from Diverse Data}
Collecting large amounts of robot data for a specific task can be challenging and time-consuming. A common strategy to overcome this limitation is to leverage data from other sources, including online robot datasets with multiple types of robots and even human videos. Prior works \cite{karamcheti_language-driven_2023}\cite{nair2022r3m}\cite{majumdar_where_2024}\cite{radosavovic2023real} train a visual encoder using both human and robot video data to avoid learning the policy network from scratch. However, the pretrained encoder only provides visual representations instead of directly learning to make decisions. Thus, to learn full policies across multiple robot embodiments, recent works have also looked at learning a single network across multiple robot embodiments by using a modular structure with a common trunk and various heads for different action spaces \cite{team2024octo}\cite{wang2024scaling}. Another common approach is the vision-language-action model (VLA) approach, where a pretrained vision-language model is finetuned with robot data to act as a robot policy \cite{brohan2023rt}\cite{black2410pi0}\cite{kim_openvla_2024}. However, due to their large size, both these approaches can be expensive to fine-tune to improve performance on a specific task, and sometimes suffer from slow inference. We compare our approach against a finetuned HPT \cite{wang2024scaling} as a representative example of these types of models.

Human video is also abundant, but can be challenging to use for robot policies as it lacks specific action information. Some works have attempted to overcome this by learning reusable priors or affordances from human video that can then accelerate robot learning~\cite{bahl_human--robot_2022}\cite{bahl_affordances_2023}\cite{mendonca2023structured}. However, these approaches tend to make specific assumptions about how humans interact with the world, and may not generalize to tasks that don't fit these assumptions.

\vspace{-0.1em}

\subsection{Policy Steering \& Planning with World Models} 

Recurrent-state-space-based world models (RSSM) have become an increasingly common approach to model environment dynamics and transition functions in robotics. Popularized by Hafner et al. \cite{hafner2019learning}\cite{hafner2023mastering}, these models consist of an encoder, which maps visual observations to a latent state representation, a latent transition function which predicts future latent states, and a decoder which is used primarily during training to propagate gradients. Once trained, these world models can be utilized in a variety of ways. One approach is to learn actor and critic models from the same training data to generate action sequences, roll them out over time, and evaluate them with the critic \cite{hafner2023mastering}. Another approach is to encode a goal image in the latent space and then use the similarity between planned states and the goal image as a value function to select the best sequence of actions. These action sequences are commonly optimized using gradient-free optimizers like the Cross-Entropy Method (CEM), but gradient-based optimization can also be used~\cite{zhou2024dino}\cite{mendonca2023structured}\cite{hafner2023mastering}.

These approaches are closely related to policy steering, where a value function and optionally a world model are used to refine the output of a base policy~\cite{nakamoto2024steering, wu2025foresight}. Policy steering has been shown to compensate for failure modes in the base policy~\cite{nakamoto2024steering} or to select actions from the base policy that obey safety constraints~\cite{wu2025foresight}. Using a world model allows the robot to steer its policy based not just on the next action, but on a projected series of actions. We build on the ideas from both planning with World Models and Policy Steering by learning a value function to steer actions from a behavior cloned policy towards states within the data distribution, and closer to the task goal.

\subsection{Offline \& Inverse Reinforcement Learning}
We propose a value function that favors states in the distribution of the training data, which is related to the pessimism used in offline RL \cite{fujimoto2019off}\cite{yu2020mopo}\cite{kidambi2020morel}\cite{kumar2020conservative}, and the concept of Inverse RL \cite{ng2000algorithms}. In offline RL, a model ( e.g., value functions, dynamics models) only has access to the state-action pairs in an offline dataset. Therefore,  predictions for state-action pairs outside of the training data distribution could be unreliable. For example, a value function could be overly optimistic when queried with out-of-distribution state-action pairs, leading to extrapolation errors \cite{fujimoto2019off}.

\subsection{Optical Flow as a Representation in Robotics}
Both 2D optical flow and 3D point flow have commonly been used as intermediate representations in robotics. Multiple works have used 3D flow to represent object affordances~\cite{eisner_flowbot3d_2024, zhang_flowbot_2024, yuan_general_2024}. These representations transfer well from simulation to real hardware and can be learned from human videos. Other works have predicted 2D optical flow as an action representation~\cite{lin_flowretrieval_2024, goyal_ifor_2022} or used optical flow to measure similarity between actions across episodes~\cite{lin_flowretrieval_2024, verghese_skills_2024}. Rather than using optical flow as a predicted action representation, in this work, we use optical flow as a unified input action representation to our world model to encode actions across diverse embodiments.

\section{Preliminaries}

We consider robot control in a Partially Observable Markov Decision Process (POMDP) setting, with a latent state space learned by a neural network. A POMDP can be described by a tuple $ ( \Omega, \mathcal{X, S, A, R, P} )$, which consists of observations $x \in  \mathcal{X}$,  conditional observation probabilities $\Omega(x' | s, a)$, states $s \in \mathcal{S}$, actions $ a \in \mathcal{A}$, reward function $r = \mathcal{R}(s, a)$, and transition function $P(s' | s, a)$. At the timestep $t \in [1,2, ..., T]$, the observation, state, action, and reward are denoted as $x_t, s_t, a_t, r_t$. Given discount factor $\gamma$, the expected future discounted reward from state $s_t$ is defined as: $E[ \sum_{t=0}^{\infty} \gamma^t r_t ]$, and can be estimated by a state-value function $\mathcal{V}(s_t)$. A dataset consists of sequences of ($x_t, a_t, r_t$) where $r_t$ is a binary reward indicating task successes. In this work, we consider two such datasets: A small dataset $\mathcal{E}$ of expert demonstrations on the target robot embodiment, and a larger cross-embodiment dataset $\mathcal{C}$ with action space $\mathcal{A}'$ that may not match the action space $\mathcal{A}$ of the target embodiment and sequences that do not necessarily represent task success. Importantly, $\|\mathcal{C}\| >> \|\mathcal{E}\|$, and we would like to maximally leverage $\mathcal{C}$ in training despite its action-space mismatches and non-optimal data. Given a horizon length $h$, policy steering seeks the best action plan $\textbf{a}_t^{*} = [a^{*}_t, a^{*}_{t+1}, ..., a^{*}_{t+h}]$ such that: $\textbf{a}_t^{*} =$ argmax $_{a \in \mathcal{A}} $  $\mathbb{E}_{ s' \sim P( \cdot | s, a)} \mathcal{R}(s, a)$.

\section{Methods}

\subsection{Flow-as-Action: Embodiment-Agnostic World Modeling}\label{section:flow-as-action}

A pretrained WM that can be easily adapted to a new embodiment should be less dependent on embodiment-specific information, such as proprioception and actions. Thus, we drop embodiment-specific information, such as proprioception, and replace embodiment-specific robot actions with an embodiment-agnostic action representation. One of our key observations is that different embodiments share similar visual motion patterns when they execute similar skills ( e.g., pick up an object), as shown in Fig.~\ref{fig:optical_flow_visualization}. Therefore, we choose optical flow, which captures visual motion and is computable via off-the-shelf tools, as the embodiment-agnostic action representation to train the WM.

Our WM adopts the objective and architecture from Dreamer v3 \cite{hafner2023mastering}, and adds an optical flow encoder during pretraining (middle of the Fig.~\ref{fig:motivation1}). The optical flow encoder is a convolution-based encoder, which predicts an $n$-dimensional vector. We set $n=\|\mathcal{A}\|$, the action space dimension of the target embodiment. Regardless of robot embodiment, the robot action space will be much smaller than the dimensionality of the optical flow image, which forces the network to extract salient features from flow and ignore noise or motion-irrelevant information (e.g., morphology differences across embodiments) when the WM is trained end-to-end on dataset $\mathcal{C}$. Setting $n$ this way also enables our simple target-embodiment finetuning procedure in the next step.

\subsection{Target Embodiment Finetuning}\label{section:finetuning}
Given our small target embodiment dataset $\mathcal{E}$, we must adapt our pretrained WM to use the target embodiment action space and steer a policy toward better actions. To this end, we first train a base policy $\pi$ with the target embodiment data (upper right of the Fig.~\ref{fig:motivation1}). This policy gives us a distribution across actions for any observation $a_t \sim \pi(x_t)$. To finetune the WM, we replace the optical flow encoder, which projected optical flow to an $\|\mathcal{A}\|$-dimensional vector, with normalized robot actions in this same $\|\mathcal{A}\|$-dimensional space (middle of the Fig.~\ref{fig:motivation1}). The WM is trained on the target embodiment dataset using the same Dreamer v3 \cite{hafner2023mastering} objective. While there is naturally a distribution shift between the compressed optical flow representation and normalized robot actions, in practice, we found this approach to be both simpler and more effective than an approach that attempted to specifically align compressed optical flow with robot actions on the limited target embodiment dataset.

Finally, we also jointly train a robust value function $\mathcal{V}(s_t)$ along with the WM, which estimates the discounted future rewards. This value function leverages the latent state representation from the WM, but its gradients do not flow back into the WM. Crucially, our value function must be able to estimate discounted future rewards across states the policy is likely to visit during inference, not just states in the expert dataset. Furthermore, the base-policy may become unstable if the robot deviates too far from states in the expert dataset, so our value function should also ensure the robot stays close to states in that expert dataset. 

To train a value function with both of these qualities, we adopt the following procedure. First, given a sequence of expert data for a horizon h, we sample a predicted action plan from the policy given the same initial observation (line 4, Alg.~\ref{alg:lps_single}), compare the resulting latent states against the latent states from the expert dataset via a similarity metric (line 6, Alg.~\ref{alg:lps_single}), and convert this similarity into and additional reward (line 8, Alg.~\ref{alg:lps_single}). We use cosine similarity given its prevalent usage in deep learning literature \cite{radford2021learning}\cite{chen2021exploring}. Thus, a robust value function is obtained by training both states likely to be visited by the policy during inference, and states from the expert dataset (line 10-11, Alg.~\ref{alg:lps_single}), given a lambda-return objective adopted from Dreamer v3 \cite{hafner2023mastering}.

\begin{algorithm}[ht]
\caption{Robust value function for LPS}\label{alg:lps_single}
\begin{algorithmic}[1]
\State \textbf{Requires:} Robot dataset $\mathcal{E}=[ (x_{1:T}, a_{1:T}, r_{1:T})_n | n=[1, ..., N]]$, world model $\mathcal{W}$, policy $\pi$, similarity metric $sim$, lambda-return computation $r_{\lambda}$, and loss $\mathcal{L}$. \\
\textbf{Initialize:} the value function $\mathcal{V}$, $\mathcal{V}_{target}$, $h$, $\gamma$, and $\lambda$.

\State Samples $(x_{t:t+h}, a_{t:t+h}, r_{t:t+h})$ from $\mathcal{E}$. 

\State Sample a predicted action plan: $ a'_{t:t+h} \sim \pi( x_{t})$.
\State $s_t = \mathcal{W}.\text{enc}(x_t)$ \Comment{  observations to latent states.}
\State $s_{t:t+h}, s'_{t:t+h} = \mathcal{W}(s_t, a_{t:t+h}), \mathcal{W}(s_t, a'_{t:t+h})$ 
\State $v_{t:t+h}, v'_{t:t+h} = \mathcal{V}_{target}(s_{t:t+h}), \mathcal{V}_{target}(s'_{t:t+h})$ 
\State $r'_{t: t + h} = r_{t: t + h} +$ $(sim(s_{t: t + h}, s'_{t: t + h})-1)/2$
 \hspace{0.5cm} \Comment{ reward penalizing deviations from the expert data}
\State $R_{\lambda}, R'_{\lambda}= r_{\lambda}(r_{t:t+h}, v_{t:t+h}, \gamma, \lambda ),r_{\lambda}(r'_{t:t+h}, v'_{t:t+h}, \gamma, \lambda )$
\State $R_{\text{all}}, S_{\text{all}}  \gets \text{concat}(R_{\lambda}, R'_{\lambda}), \text{concat}(s_{t:t+h}, s'_{t:t+h}) $
\State Optimize the value function: $\mathcal{L}( R_{\text{all}}, \mathcal{V}(S_{\text{all}}))$.
\State Updates $\mathcal{V}_{target}$ given $\mathcal{V}$.
\end{algorithmic}
\end{algorithm}

\subsection{Latent Policy Steering}

We refer to our method of finetuning a WM and training a robust value function steering a policy back to the expert dataset via WM-generated latent states as Latent Policy Steering (LPS). LPS improves the base behavior cloning policy during inference by evaluating multiple candidates in the WM's latent state space and selecting the best one according to the value function. We sample $B$ number of action plans with horizon $h$. After encoding the current observation, the WM predicts the corresponding future states for each plan and computes the state values. A plan-level value is computed based on a weighted average of state values with a discount factor, by assigning heavier weights to future states to reduce noise in the value function predictions. In practice, we found that this discounted weighted averaging of values works better than using the average or the last-state value per plan to represent the plan-level value. Finally, the action plan with the highest value is executed for the given horizon before replanning.

\begin{figure}[t]
      \centering
     
      \includegraphics[scale=0.95]{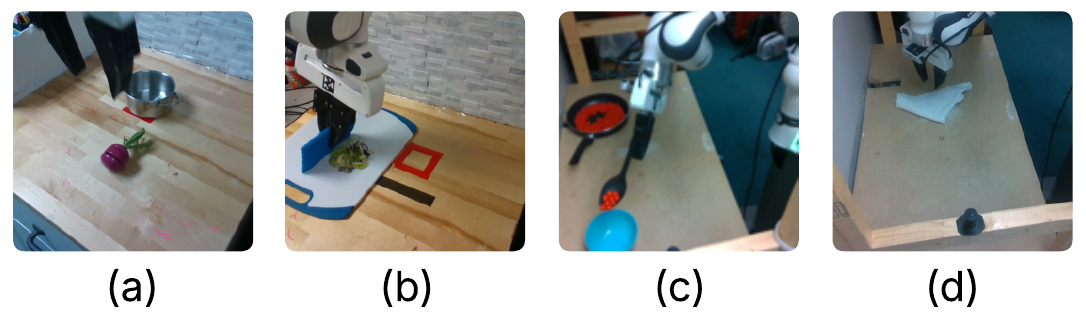}
      \caption{ 
      \textbf{Real-world experiment}. The proposed method and baselines are evaluated on: (a) Put-radish-in-pot, (b) Sweep-salad-off-the-board, (c) Scoop-beads-with-spoon, and (d) Fold-towel-to-triangle. }
      \vspace{-1em}
      \label{fig:realworld}
   \end{figure}
   
\section{Experiments}

We evaluate the proposed method on real-world environments and on a simulated benchmark (Robomimic \cite{robomimic2021}). We compared our embodiment-agnostic approach to baselines with an embodiment-dependent pretrained policy \cite{wang2024scaling} and learning from scratch. Ablations are provided to verify the key choices of the algorithm and the effectiveness of the embodiment-agnostic actions across multiple embodiments. In the following, we first describe our experiment settings, including environment, datasets, and baselines.

\subsection{Settings}

\textbf{Environment: Real world.} We create an evaluation environment with a Franka robot, a desk, and objects to manipulate. We mount a camera on the robot's wrist and a fixed camera on the side. 
The target dataset is collected via teleoperation with a joystick. The positions of the objects during data collection are randomly selected from the desk area. The object positions during evaluations include 20 positions that are pre-selected, uniformly covering the desk area. We consider a pick-and-place task and three challenging long-horizon tasks that require tool use or precise manipulations of deformable objects (sweep-salad-off-the-board, scoop-beads-with-spoon, fold-towel-to-triangle), shown in Fig.~\ref{fig:realworld}. Scooping requires the robot to use a spoon to move small beads from a pan to a bowl. Imprecise manipulation will easily lead to beads falling out of the spoon. Folding/Sweeping both require interactions to manipulate deformable objects.

\textbf{Environment: Robomimic.} Our approach and baselines are evaluated on a Franka robot across 4 tasks (Lift, Can, Square, Transport) from Robomimic~\cite{robomimic2021}. We only use the 30-50 expert demonstrations to learn/finetune our models, mimicking a low-data regime setting. We use Robosuite~\cite{robosuite2020} with a joystick to collect 100 demonstrations on 3 robots other than Franka: IIWA, UR5e, and Kinova3 via the provided interface. In this case, the pretraining data has action labels inherently different from those of the Franka robot. Our ablation study for LPS's design choices and different horizons uses 100 demonstrations. Our evaluation protocol is similar to \cite{ren2024diffusionpolicypolicyoptimization}. We report the average success rates and standard deviations across 3 seeds. Each seed has a success rate based on hundreds of evaluations.

\textbf{Datasets.} We use pretraining datasets $\mathcal{C}$ and target datasets $\mathcal{E}$ in our experiment. The target dataset is a small expert quality (all demonstrations succeed and result in task completions) Franka dataset with robot actions (e.g., end-effector pose), which is used to learn a policy and finetune the WM. The pretrain dataset $\mathcal{C}$ has multiple embodiments, with no assumptions on quality. The pretraining dataset is composed of data collected in simulation, existing public robot datasets, or easy-to-collect human data. In our case, we ask the human to interact with the real-world environment without a specific purpose, known as data from play. We consider 3 pretraining data variants: 
\begin{itemize}
    \item -sim: A dataset of simulation robot data collected by a human teleoperator. It includes 3 robots, different from the robot used for evaluation (target robot), with 100 demonstrations for each robot. 
    \item -real: A set of public robot data collected by different research groups in the real world ( Open X-Embodiment \cite{vuong2023open}). Since most of the object manipulation data comes from several popular robots in the Open X-embodiment dataset, we filter data from 4 popular robots. The resulting dataset has 2000 episodes after dropping episodes that are too short or too long.
    \item -human: A dataset made of human video data from play. A human is asked to interact with the environment without specific goals. Therefore, frequent unintentional task completions/resets throughout the playing episode contain richer information to learn, compared to an episode collected via teleoperations. Its total transitions are about 43\% of the public robot dataset.
    \item -mix: the mixture of robot data in simulation, real world,  and human data. In total, it has 8 different embodiments, including 3 robots in simulation, 4 robots in the real world, and a human embodiment.
\end{itemize}

\textbf{Baselines.} Variants of our approach are denoted by \textbf{*}. The suffix of LPS (-sim/real/human/mix) denotes the variants of our approach, with different pretraining dataset mixtures.
\begin{itemize}
    \item BC: a diffusion policy \cite{chi2023diffusion} learned from scratch via behavior cloning on the target dataset. All of our experiments use an action prediction horizon of 16.
    \item HPT: a cross-embodiment pretrained policy \cite{wang2024scaling} finetuned on the target dataset, including embodiment-specific action head and the low-level encoder. It uses an action prediction horizon of 16.
    \item LPS*: our proposed method without pretraining the WM. The WM is trained only on the target dataset. The WM will be combined with the policy (BC), and has a prediction horizon of 16. The best action plan, according to the WM, is executed in an open loop.
    \item LPS-(sim/real/human/mix)*: LPS with WMs pretrained on robot data from simulations (-sim), public robot data collected in the real world (-real), human videos collected in the real world(-human), or the mixture of -sim/real/human. It has a prediction horizon of 16.
\end{itemize} 
LPS* and LPS-(sim/real/human/mix)* use the same base policy as the BC baseline. Except for HPT, all policies are learned from scratch, given a small target dataset.

\subsection{Real-world experiments}\label{section:real}

\begin{table}[t]
\caption{Latent Policy Steering in the real world}

\label{table:real}
\vspace{0.5em}
\setlength{\tabcolsep}{0.5pt}   
\renewcommand{\arraystretch}{0.95}
\begin{tabular}{@{}lcccccccc@{}}
\toprule
\multirow{3}{*}{Task\textbackslash{}Settings} & \multicolumn{4}{c}{ 30$^a$ or 50$^b$ demonstrations} & \multicolumn{4}{c}{ \shortstack{60$^c$ or 100$^d$ demonstrations}
} \\ \cmidrule(l){2-9} 
 & \multicolumn{2}{c}{\shortstack{From\\scratch}} & \multicolumn{2}{c}{\shortstack{Pretrain-\\and-finetune}} & \multicolumn{2}{c}{\shortstack{From\\scratch}} & \multicolumn{2}{c}{\shortstack{Pretrain-\\and-finetune}} \\
 & BC & LPS* & HPT &  \begin{tabular}[c]{@{}c@{}}LPS-mix*\end{tabular} & BC & LPS* & HPT  & \begin{tabular}[c]{@{}c@{}}LPS-mix*\end{tabular} \\ \cmidrule(r){1-1}

\begin{tabular}[c]{@{}l@{}}Put-radish-\\ in-pot\end{tabular} & 7/20$^{^a}$ & 8/20$^{^a}$ & 4/20$^{^a}$ &  \textbf{11/20}$^{^a}$ & 13/20$^{^c}$ & 15/20$^{^c}$ & 6/20$^{^c}$ & \textbf{19/20}$^{^c}$ \\
\begin{tabular}[c]{@{}l@{}}Sweep-salad-\\ off-the-board\end{tabular} & 4/20$^{^a}$ & 4/20$^{^a}$ & 0/20$^{^a}$ &  \textbf{6/20}$^{^a}$ & 6/20$^{^c}$ & 8/20$^{^c}$ & 2/20$^{^c}$ &\textbf{11/20}$^{^c}$ \\
\begin{tabular}[c]{@{}l@{}}Scoop-beads-\\ with-spoon\end{tabular} & 6/20$^{^b}$ & 5/20$^{^b}$ & 0/20$^{^b}$ & \textbf{10/20}$^{^b}$ & 13/20$^{^d}$ & \textbf{16/20$^{^d}$} & 0/20$^{^d}$ & \textbf{16/20}$^{^d}$ \\
\begin{tabular}[c]{@{}l@{}}Fold-towel-\\ to-triangle\end{tabular} & 0/20$^{^b}$ & 0/20$^{^b}$ & 0/20$^{^b}$ & \textbf{2/20}$^{^b}$ & 9/20$^d$ & 11/20$^{^d}$  & 3/20$^d$  & \textbf{13/20}$^d$  \\ 

\begin{tabular}[c]{@{}l@{}}Average\end{tabular} & 21.2\% & 21.2\% & 5.0\%  & \textbf{36.2}\%& 51.2\% & 62.5\% & 13.8\% & \textbf{73.8}\% \\

\bottomrule
\end{tabular}

\vspace{0.5em}
We report the number of successes out of 20 trials in the real world. Thanks to the embodiment-agnostic pretrained WM, LPS-mix* has  significantly improved the base policy (BC) performance across tasks. Although HPT was pretrained on a large-scale dataset with more than 20 embodiments, such an embodiment-dependent pretrained policy performs poorly given a small amount of target-embodiment data for finetuning.

\vspace{-1.5em}
\end{table}

\begin{table*}[t]
\caption{Latent Policy Steering in Robomimic}
\label{table:robomimic}
\vspace{-0.5em}
\begin{minipage}{0.8\linewidth}
\begin{tabular}{@{}llcccccccccccc@{}}
\toprule
\multirow{7}{*}  & \multirow{3}{*}{ \shortstack[1]{Task\textbackslash\\Setting}}                               & \multicolumn{6}{c}{30 Franka demonstration}                                                                                                                                                                                                            & \multicolumn{6}{c}{50 Franka demonstration}                                                                                                                                                                                                            \\ \cmidrule(l){3-14} 
                            &                                                                               & \multicolumn{2}{c}{From scratch} & \multicolumn{4}{c}{Pretrain-and-finetune}                                                                                                                                                                           & \multicolumn{2}{c}{From scratch} & \multicolumn{4}{c}{Pretrain-and-finetune}                                                                                                                                                                           \\
                            &                                                                               & BC              & LPS*        & \multicolumn{2}{c}{\begin{tabular}[c]{@{}c@{}}LPS\\-sim*\end{tabular}} & \begin{tabular}[c]{@{}c@{}}LPS\\-real*\end{tabular}  & \begin{tabular}[c]{@{}c@{}}LPS\\-mix*\end{tabular} & BC              & LPS*        & \multicolumn{2}{c}{\begin{tabular}[c]{@{}c@{}}LPS\\-sim*\end{tabular}} & \begin{tabular}[c]{@{}c@{}}LPS\\-real*\end{tabular}  & \begin{tabular}[c]{@{}c@{}}LPS-\\mix*\end{tabular} \\ \cmidrule(lr){2-2}
                            & Lift                                                                          & \textbf{25.6}±3.6      & 18.0±5.6     & \multicolumn{2}{c}{24.7±1.4}                                                    & 24.5±1.7                                                             & 24.4±1.9                                                             & 82.0±6.2      & 52.7±5.0     & \multicolumn{2}{c}{\textbf{84.4}±10.8}                                                   & 84.0±1.5                                                               & \textbf{84.4}±10.8                                                             \\
                            & Can                                                                           & 64.9±10.2     & \textbf{74.7}±3.1     & \multicolumn{2}{c}{71.1±1.0}                                                    & 71.5±1.7                                                              & 73.1±1.5                                                             & 76.7±2.1      & 80.2±4.9     & \multicolumn{2}{c}{84.9±3.6}                                                    & 85.3±3.8                                                      & \textbf{85.8}±4.1                                                    \\
                            & Square                                                                        & 19.4±2.5      & 20.6±3.6     & \multicolumn{2}{c}{\textbf{24.2}±5.6}                                                    & 21.3±3.6                                                               & 21.3±3.6                                                             & 44.8±6.5      & 48.3±19.7    & \multicolumn{2}{c}{\textbf{52.4}±7.2}                                                    &  49.5±0.6                                                               & 49.0±6.4                                                             \\
                            & Transport                                                                        & 22.8±2.1      & 22.5±4.1     & \multicolumn{2}{c}{\textbf{23.5}±0.7}                                                    & 19.8±1.2                                                               & 23.2±0.5                                                             & 25.8±1.6      & 30.4±3.1    & \multicolumn{2}{c}{27.9±6.2}                                                    &  32.9±2.8                                                               & \textbf{34.6}±3.6                                                             \\
                            & Average                                                                       & 33.2\%          & 34.0\%         & \multicolumn{2}{c}{\textbf{35.9}\%}                                                        & 34.3\%                                                               & 35.5\%                                                             & 57.3\%          & 52.9\%         & \multicolumn{2}{c}{62.4\%}                                                        & 62.9\%                                                               & \textbf{63.4}\%                                                             
                                                 \\ \bottomrule
\end{tabular}
\end{minipage}
\vspace{0.5em}
\\We report the mean success rate across 3 seeds with standard deviations. Despite the majority of the pretrain data are collected in the real-world, our proposed method (e.g., LPS-mix*) achieves 10.6\% relative improvement based on the behavior cloned policy in the simulation, 50 demonstrations, averaged across 4 tasks. Since transport is a long-horizon bimanual task with a greater action dimension and requiring precision, the LPS's significant improvement over the baseline (34\% relative improvement) with sufficient demonstrations shows its potential beyond simple tasks and single-arm robot systems. The improvement from LPS with or without pretraining is less significant, given 30 demonstrations. We believe LPS performs best when given a multimodal policy (i.e., a policy that generates diverse plans), whereas 30 demonstrations could lead to unimodal policies that are difficult to improve.
\vspace{-2em}
\end{table*}

\textbf{ Does the embodiment-agnostic pretrained WM leverage multi-embodiment data more effectively than an embodiment-dependent pretrained policy?} We evaluate LPS-mix* against three other baselines (BC, LPS*, HPT) in the real world on a pick-and-place task and three challenging tasks involving tool-use, and deformable object manipulation. LPS-mix* significantly improves the policy (BC) performance across tasks (Table~\ref{table:real}) (relative improvement of 70\% and 44\%, given 30-50 and 60-100 demonstrations). While HPT has been pretrained on a large-scale dataset with more than 20 embodiments, its embodiment-dependent pretraining makes it challenging to finetune to a target-embodiment in a low-data regime. On the other hand, by using an embodiment-agnostic pretrained WM, LPS can translate an extra multi-embodiment dataset into better performance on an unseen target embodiment during pretraining, with only a small amount of data. Both BC and LPS* that learn from scratch perform poorly, possibly due to a lack of data. 

\subsection{Ablation studies in the simulated benchmark}

\begin{table}[b]
\caption{WMs pretrain with optical flow vs. EEF actions}\label{table:flowandeef}
\vspace{-0.5em}

\begin{minipage}{0.8\linewidth}

\begin{tabular}{@{}lllll@{}}
\toprule
\multirow{2}{*}{Task\textbackslash{}Settings} & \multicolumn{2}{c}{30 demonstrations} & \multicolumn{2}{c}{50 demonstrations} \\ \cmidrule(l){2-5} 
 & \begin{tabular}[c]{@{}c@{}}LPS-sim\\(flow)*\end{tabular} & \shortstack{LPS-sim\\(eef)} & \begin{tabular}[c]{@{}c@{}}LPS-sim\\(flow)*\end{tabular} & \shortstack{LPS-sim\\(eef)} \\ \cmidrule(r){1-1}
Lift & \textbf{24.7}±1.4 & 24.6±2.0 & \textbf{84.4}±10.8 & 81.9±8.7 \\
Can & 71.1±1.0 & \textbf{73.7}±17 & \textbf{84.9}±3.6 & 84.4±7.2 \\
Square & \textbf{24.2}±5.6 & 21.0±16 & \textbf{52.4}±7.2 & 45.3±21 \\
Transport & \textbf{23.5}±0.7 & 18.3±2.6 & \textbf{27.9}±6.2 & 24.6±6.4 \\
Average & \textbf{35.9\%} & 34.4\% & \textbf{62.4\%} & 59.1\% \\ \bottomrule
\end{tabular}
\end{minipage}
\vspace{0.5em}
\\We compare the WM pretrained with embodiment-agnostic actions (optical flow) against the WM pretrained with end-effector action (eef). The optical-flow-based WM produces better results on a dataset of 3 robots. With more embodiments available, we believe the performance gap will widen.

\end{table}

\begin{figure}[b]
  \centering
 
  \includegraphics[scale=0.15]{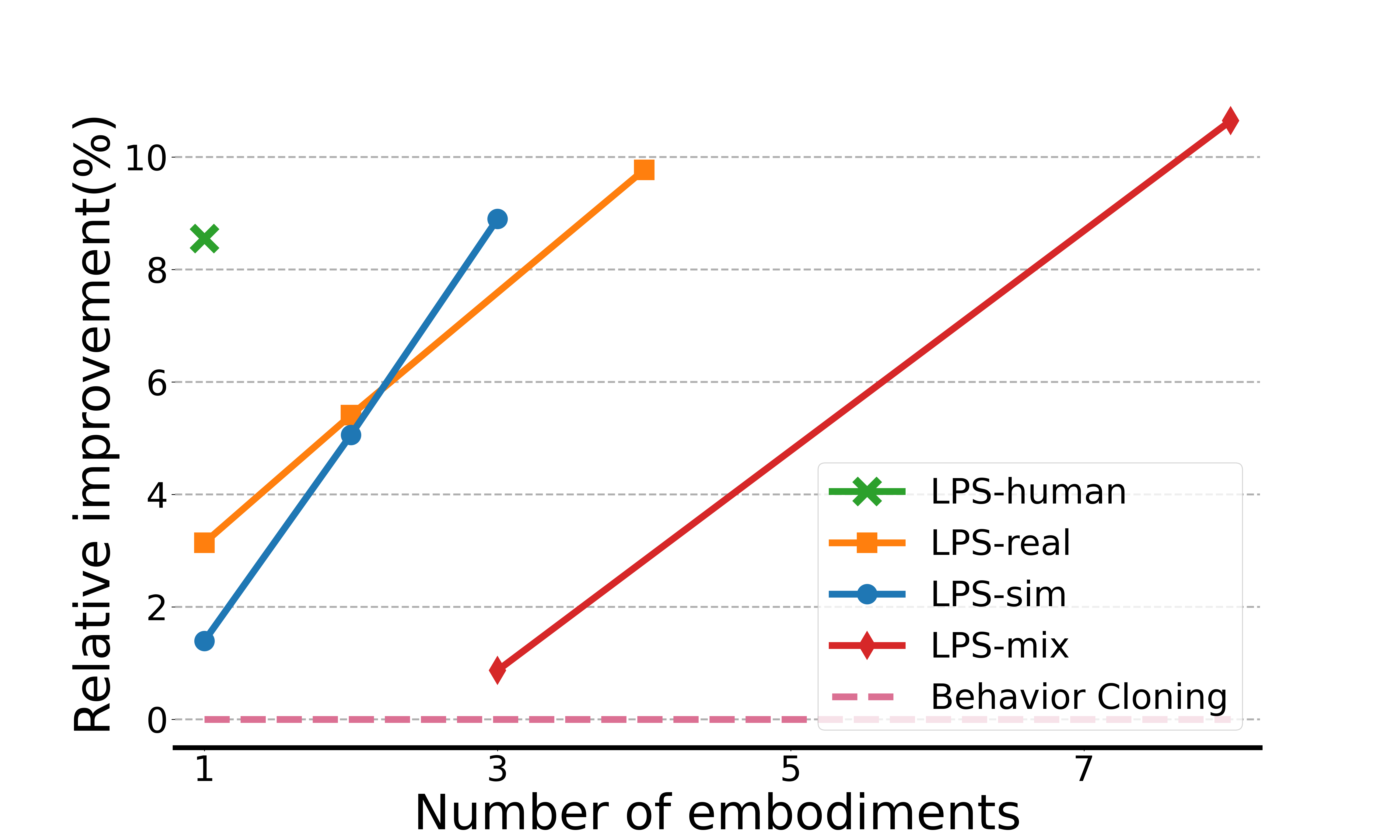}
  \caption{ 
  \textbf{Scaling through embodiments}. We evaluate LPS in the Robomimic given different pretrain mixtures. LPS-mix manages to improve the behavior cloning policies' performance up to 10.6\%, given 50 target-embodiment demonstrations, averaged across four tasks, considering simulation, real-world robot data, and human videos from play.  }
  \vspace{-1em}
  \label{fig:scaling}
\end{figure}

\textbf{
How does LPS’s performance vary with different embodiments used during pretraining?
} Given four Robomimic tasks, we compare LPS-mix* against LPS variants pretrained with real-world robot data (LPS-real*) or simulation data (LPS-sim*), and baselines without pretraining (BC, LPS*), given 30-50 demonstrations. In simulations,  we expect the performance improvement due to the embodiment-agnostic pretrained WM to be lower than in real-world experiments, given that the pretrain mixture of LPS-real* and LPS-mix* contain primarily real-world data.

Given 50 demonstrations, LPS-mix* outperforms other LPS variants and improves a behavior cloned policy by 10.6\%, despite the majority of its pretrain data being collected in the real world (robot and human embodiments). LPS's results regarding the transport task are particularly encouraging, since it is a long-horizon bimanual task with a higher action dimension (14) that requires coordination between two arms during transport. It demonstrates LPS's potential beyond simple tasks and single-arm robot systems.

LPS's improvement over the BC baseline is less significant in the 30 demonstration cases, regardless of pretraining mixture. In our opinion, LPS works best when given a multimodal policy that generates diverse plans. Too few demonstrations will result in a unimodal policy that generates similar plans and gives little room for improvement.

\textbf{How does LPS’s performance scale with the number of embodiments?} We explore how different combinations/numbers of embodiments for pretraining affect the performance, averaged across four Robomimic tasks, as shown in Fig.~\ref{fig:scaling}. Our target dataset has 50 demonstrations, and the pretraining dataset could include simulation robot data, real-world robot data, and human video data from play. We observe a promising trend: thanks to the embodiment-agnostic action representation and WM, increasing the number of embodiments leads to more usable pretraining data and results in better performance compared to the behavior-cloned baseline. While embodiment-dependent pretrained policies such as HPT \cite{wang2024scaling} could also scale with embodiments, we do not compare to it since HPT pretrained with more than 20 embodiments performs sub-optimally given a small amount of finetuning data on the target embodiment, as shown in real-world experiments (Table~\ref{table:real}).

Surprisingly, we observe that the human video data from play leads to a very competitive pretrained WM, despite having less data compared to other data mixtures with more embodiments. Our hypothesis is that humans are able to demonstrate more diverse and cohesive manipulation skills with their hands in a shorter period of time than when teleoperating a robot, resulting in higher data quality.

\textbf{How does the optical flow compare to other off-the-shelf action representations?} Robot end-effector pose (EEF) is another off-the-shelf action representation for manipulators that can be moderately embodiment-agnostic. In simulation, we investigate the advantage of using optical flow as an off-the-shelf embodiment-agnostic action representation during pretraining over EEF pose, given a pretraining dataset with 3 different robots. Our approach (LPS-sim(flow)*) uses optical flow action during pretraining, and LPS-sim(eef) uses EEF action during pretraining. Both are finetuned on the target dataset with 30 or 50 demonstrations, using the EEF robot action. 

The results are shown in Table~\ref{table:flowandeef}. The advantage of optical flow as embodiment-agnostic actions vs. EEF actions is clear, given 3 embodiments. We suspect the advantage will keep growing when the dataset size is larger and the number of embodiments is larger. Notice that optical flow also allows human embodiment to be part of pretraining, while EEF actions are only available for robots.

\begin{table}[t]
\caption{Latent Policy Steering's value function ablations}
\label{table:ablation}
\vspace{-0.5em}
\begin{minipage}{0.78\linewidth}
\begin{tabular}{@{}lcccc@{}}
\toprule
\shortstack{Task\textbackslash\\Settings} & \multicolumn{1}{l}{BC} & \multicolumn{1}{l}{ \shortstack{LPS-\\vanilla}} & \multicolumn{1}{l}{\shortstack{LPS-\\bootstrap}} & \multicolumn{1}{l}{ LPS} \\ \midrule
Lift                        & 99.1±0.4                           & 96.0±4.1                                      & 98.7±0.3                                          & \textbf{99.3} ±0.9                     \\
Can                          & 79.8±1.6                          & 82.2±4.8                                      & 80.2±4.3                                         & \textbf{87.3}±3.9                     \\
Square                       & 45.6±1.1                          & 52.0±7.1                                     & 48.5±18.8                                       &  \textbf{52.9}±9.8                  \\
Transport                    & 27.3±2.1                          & 30.5±4.5                                      & 29.6±2.9                                       & \textbf{35.3}±6.4                  \\ \midrule
Average                      & 62.9\%                          & 65.2\%                                      & 64.3\%                                         & \textbf{68.7\%}                      \\ \bottomrule
\end{tabular}
\end{minipage}
\vspace{0.5em}
\\We ablate our designs of the robust value function for LPS into 2 variants (LPS-vanilla/bootstrap). In both variants, performance is lower than the policy-only baseline (BC), demonstrating the necessity of designing a robust value function to let WM benefit the policy (BC).
\vspace{-2.0em}
\end{table}

\textbf{How does the robust value function affect LPS’s performance?} LPS uses a robust value function to steer the policy back to the expert dataset. Such a value function is trained on data from the expert dataset and the data likely to be visited by the policy during inference. We ablate these designs through 2 variants below.

\begin{itemize}
    \item LPS-vanilla: This variant only trains the value function on the data from the expert dataset, given environmental rewards. It will select the best candidate based on the value function.  
    \item LPS-bootstrap: This variant learns a value function both on the expert dataset and the data likely to be visited by the policy during inference. It achieves this by simulating the states likely visited by the policy through the WM. Different from the LPS, the extra states are labeled with the same rewards as the dataset's states and do not include the reward penalizing distribution shifts, modifying line 8 of Alg.~\ref{alg:lps_single} ($r'_{t:t+h} = r_{t:t+h}$). It will select the best candidate based on the value function.
\end{itemize}
All models use 100 demonstrations to train from scratch and no pretraining is involved.

The result shown in Table~\ref{table:ablation} demonstrates that all the designs of the value function contribute to the success of LPS. A vanilla LPS has never seen out-of-distribution data. Thus, it is not robust to the distribution shift during inference. Despite seeing out-of-distribution data, LPS-bootstrap treats it optimistically and does not penalize behavior that deviates from the training dataset. Both variants perform worse than a simple policy-only baseline (BC). This shows that a robust value that has seen in/out-of-distribution data and penalizes deviation from the dataset is crucial to make a WM helpful for policy inference.

\begin{figure}[t]
  \centering
  \includegraphics[width=0.9\columnwidth]{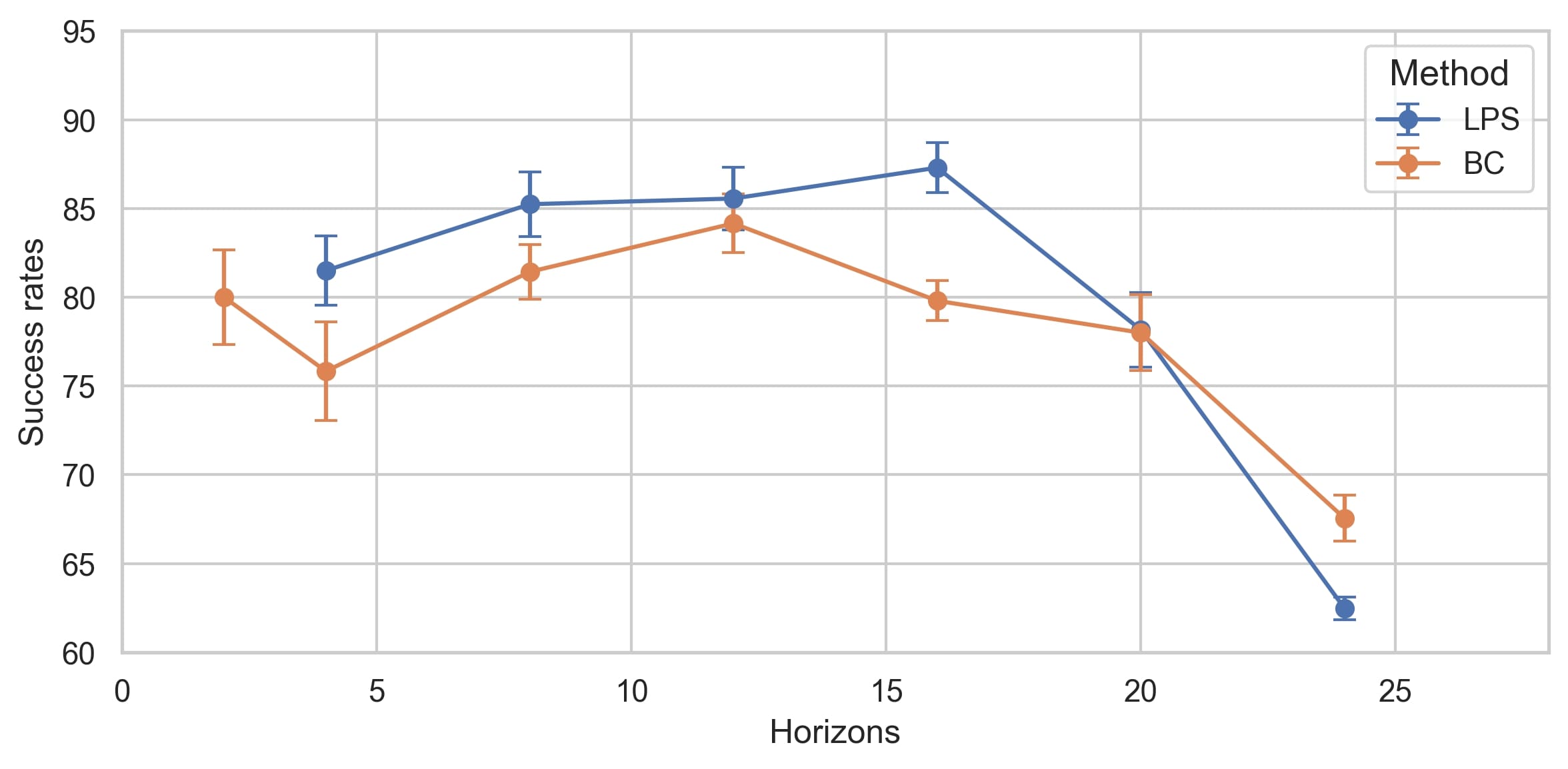}
  \caption{ \textbf{Latent Policy Steering with different horizons.} We found that the proposed LPS method can effectively improve policy performance from a shorter horizon to a longer horizon up to 16, both learned from scratch. When the horizon becomes very long (e.g., 20, 24), the reward used to capture distribution shift becomes noisy, making the value function less useful during inference. }\label{fig:lps_horizon}
  \vspace{-2em}
\end{figure}

\textbf{How does the inference horizon affect LPS’s performance?} We investigate LPS's performance given different action prediction horizons from small to large. The WM's prediction horizon always matches the action prediction horizon of the base diffusion policy \cite{chi2023diffusion}. We learn the policy and the WM from scratch, given EEF action, with 100 demonstrations on the Can task of Robomimic \cite{robomimic2021}.

We found that LPS with a WM performs better than BC at horizons 4,8,16,20, and performs worse than BC for a horizon of 24 (Fig.~\ref{fig:lps_horizon}). Large action prediction size/horizon of 24 leads to a suboptimal policy with poor performance, and the penalization on deviation from the dataset will be noisy (lines 8-10, Alg.~\ref{alg:lps_single}). This results in a noisy reward and makes the value function less useful.

\section{CONCLUSIONS \& DISCUSSION}

We pretrain an embodiment-agnostic World Model (WM) on existing or cost-effective data sources such as public multi-embodiment robot datasets \cite{vuong2023open}, robot data from simulations \cite{robosuite2020}, and easily collected human data from play to improve visuomotor policies with a small amount of real-world data. We propose optical flow as an embodiment-agnostic action representation to enable this training across diverse embodiments. The resulting pretrained WM has minimal dependency on specific embodiments and can be easily finetuned to a target embodiment with a small amount of data. Our proposed algorithm, Latent Policy Steering, manages to improve a policy's performance in the real-world evaluations across 4 tasks relatively by 70\% with 30-50 expert demonstrations, and 44\% with 60-100 expert demonstrations, showing its effectiveness over prior work without pretraining a WM.

Despite the advantages of being embodiment-agnostic and easily computable, we observe several limitations with optical flow as an action representation. Firstly, optical flow cannot reliably capture motions if an occlusion occurs. Furthermore, optical flow depends on the viewpoint. The same skill will lead to different optical flow patterns given different viewpoints. However, this limitation can be overcome by leveraging data with diverse viewpoints, such as that present in large cross-embodiment datasets. In addition, visual observations from non-static cameras (e.g., mobile robots) can produce noisy optical flow without further processing. We plan to investigate how to build more complete and scalable action representations by bridging different embodiments with embodiment-agnostic action and augmenting it with embodiment-specific details when the previous action representation lacks critical information for actions.

\bibliographystyle{IEEEtran}
\bibliography{root}

\end{document}